\documentclass[conference]{IEEEtran}
\IEEEoverridecommandlockouts

\usepackage{amsmath,amssymb,amsfonts}
\usepackage{algorithmic}
\usepackage{graphicx}
\usepackage{textcomp}
\usepackage{xcolor}
\usepackage{url}
\usepackage{multirow}
\def\BibTeX{{\rm B\kern-.05em{\sc i\kern-.025em b}\kern-.08em
    T\kern-.1667em\lower.7ex\hbox{E}\kern-.125emX}}
\begin{document}

\title{RKEFino1: A Regulation Knowledge-Enhanced Large Language Model}

\author{\IEEEauthorblockN{1\textsuperscript{st} Yan Wang*}
\IEEEauthorblockA{\textit{Yale University} \\
New Haven, CT, USA \\
yan.wang.yw937@yale.edu}
\and
\IEEEauthorblockN{2\textsuperscript{nd} Yueru He}
\IEEEauthorblockA{\textit{Columbia University} \\
New York, NY, USA \\
yh3507@columbia.edu}
\and
\IEEEauthorblockN{3\textsuperscript{rd} Ruoyu Xiang}
\IEEEauthorblockA{\textit{New York University} \\
New York, NY, USA \\
rx2306@nyu.edu}
\and
\IEEEauthorblockN{4\textsuperscript{th} Jeff Zhao}
\IEEEauthorblockA{\textit{The University of Texas at Austin}\\
Austin, Texas, USA \\
jeffzhao@utexas.edu}
}

\maketitle

\begin{abstract}
Recent advances in large language models (LLMs) hold great promise for financial applications but introduce critical accuracy and compliance challenges in Digital Regulatory Reporting (DRR). To address these issues, we propose RKEFino1, a regulation knowledge-enhanced financial reasoning model built upon Fino1, fine-tuned with domain knowledge from XBRL, CDM, and MOF. We formulate two QA tasks—knowledge-based and mathematical reasoning—and introduce a novel Numerical NER task covering financial entities in both sentences and tables. Experimental results demonstrate the effectiveness and generalization capacity of RKEFino1 in compliance-critical financial tasks. We have released our model on Hugging Face\footnote{\url{https://huggingface.co/YanAdjeNole/RKEFino1-14B}}.
\end{abstract}


\section{Introduction}
The financial industry increasingly leverages reinforcement learning (RL) techniques, giving rise to the interdisciplinary field known as Financial Reinforcement Learning (FinRL), which includes applications in portfolio management, algorithmic trading, and option pricing. Recent advancements in large language models (LLMs) have further accelerated the growth of open finance by providing scalable, affordable, and personalized solutions such as enhanced financial search and robo-advisory services. 

Despite these promising developments, the application of LLMs to Digital Regulatory Reporting (DRR) introduces significant challenges due to the complexity and precision required by financial regulations. Financial reporting standards such as the eXtensible Business Reporting Language (XBRL) commonly experience high error rates, highlighting critical vulnerabilities in existing reporting mechanisms. Additionally, managing financial transactions through the Common Domain Model (CDM) demands accurate lifecycle tracking, while the Model Openness Framework (MOF) requires comprehensive transparency and completeness from machine learning models. Issues with misinformation and inaccuracies—hallucinations—that plague current LLMs pose severe risks, potentially leading to regulatory non-compliance, financial losses, and diminished trust among stakeholders. 

To address these critical challenges, we introduce an RKEFino1, a regulation knowledge-enhanced Fino1 model~\cite{qian2025fino1transferabilityreasoningenhanced}, fine-tuned specifically with domain knowledge from XBRL, CDM, and MOF. This enhanced model aims to significantly improve interpretability, compliance accuracy, and reliability in digital regulatory reporting tasks, thereby contributing effectively to market integrity and regulatory compliance.

\section{Related Works}

Large Language Models (LLMs) have demonstrated significant success across various tasks in the general domain, such as information extraction~\cite{keloth2024advancing}, retrieval~\cite{wang2024cdemapper}, ranking~\cite{wang2025ordrankben}, and summarization~\cite{zhang2024comprehensive}. Inspired by these promising outcomes, recent research has progressively extended the application of LLMs into finance-specific scenarios. Initial efforts have focused on adapting existing models for financial applications, demonstrating their capabilities in financial text analysis~\cite{yang2020finbert}, market sentiment prediction~\cite{yang2023fingpt}, and automated trading strategies~\cite{kou2024automate}. Building upon these findings, recent advancements have led to the development of specialized financial LLMs, including BloombergGPT~\cite{wu2023bloomberggpt}, FinGPT~\cite{yang2023fingpt}, PIXIU~\cite{xie2023pixiu}, OpenFinLLMs~\cite{xie2024open}, and Plutus~\cite{peng2025plutusbenchmarkinglargelanguage}. These finance-specific models leverage pretraining on financial corpora, significantly enhancing their ability to understand and accurately interpret domain-specific terminology and contexts. Empirical evaluations confirm that these adapted LLMs effectively handle complex financial tasks, such as financial market sentiment analysis, earnings call analysis, regulatory compliance monitoring, fraud detection, portfolio optimization, and risk management.

\section{Methodology}
In this section, we primarily focus on introducing our RKEFino1 model, defining our fine-tuning tasks, collecting training data, and setting model parameters.

\subsection{RKEFino1: regulation knowledge-enhanced Fino1}
Fino1~\cite{qian2025fino1transferabilityreasoningenhanced} is a lightweight financial reasoning model built upon LLaMA-3.1-8B-Instruct, designed for accurate and efficient deployment in real-world financial scenarios. It adopts a two-stage training strategy combining supervised fine-tuning and reinforcement learning, guided by curated reasoning paths.
In the first stage, Fino1 learns domain-specific reasoning patterns from financial question-answer pairs, with an emphasis on step-by-step interpretability. In the second stage, the model leverages verifier-guided reinforcement learning to iteratively refine its outputs, enhancing correctness and coherence through techniques such as backtracking, verification, and correction.

This design enables Fino1 to achieve strong performance on financial mathematical reasoning tasks. However, it cannot analyze Digital Regulatory Reporting (DRR) scenarios, which require a nuanced understanding of regulatory structures and compliance logic. To address this limitation, we further fine-tune Fino1 on curated DRR-related datasets, incorporating domain-specific knowledge from financial regulations. The resulting model, RKEFino1 (Regulation Knowledge-Enhanced Fino1), is equipped with enhanced reasoning abilities tailored to regulatory reporting and compliance tasks.

\subsection{Task formulation}\label{task_formulation}
To enhance regulatory knowledge in the Fino1 model, we define a series of Digital Regulatory Reporting (DRR) question-answering tasks based on CDM, MOF, and XBRL data, specifically categorized into knowledge-based QA and mathematical reasoning QA. For the knowledge-based QA task, given a regulation-related question $q$, covering licenses, abbreviations, terminologies, and tags. The model generates answers $a$ by determining license applicability, expanding abbreviations, explaining terminologies, and identifying tags based on its inherent knowledge.

\begin{equation}
    a = f(q)
\end{equation}

For the mathematical reasoning QA task, given a mathematical question $q$ about a financial report, a financial formulation $t$, and descriptions $d$ of terms in the formulation, the model calculates a numerical answer $a$ by leveraging its comprehensive regulatory knowledge, financial understanding, and inference capabilities. So,

\begin{equation}
    a = f(q,t,d)
\end{equation}

\subsection{Data collection}
We collect our training data mainly from 3 data source: the CDM documentations\footnote{\url{https://cdm.finos.org/}} for CDM knowledge, the Open Source Initiative (OSI) website\footnote{\url{https://opensource.org/licenses}} for MOF knowledge, and the U.S. Securities and Exchange Commission (SEC) website\footnote{\url{https://www.sec.gov/}} for XBRL knowledge. Moreover, we also collect extra training data from existing datasets such as  XBRL terminology\footnote{\url{https://huggingface.co/datasets/KirkHan/XBRL_Terminology}}. Therefore, as shown in Table~\ref{tab_1}, we collected 9,898 training samples for our model.

\begin{table}[htbp]
\caption{The statistic of training data.}
\begin{center}
\begin{tabular}{|c|c|c|c|}
\hline
\textbf{Task} & \textbf{Domain}& \textbf{\#Samples}& \textbf{Source} \\
\hline
\multirow{3}*{K-QA$^{\mathrm{a}}$} & CDM& 478& CDM documentation \\
~& MOF& 258& OSI website \\
~ & XBRL& 8,052& SEC and XBRL Terminology \\
\hline
MR-QA$^{\mathrm{b}}$ &XBRL & 1,110 & SEC \\
\hline
\multicolumn{4}{l}{$^{\mathrm{a}}$ K-QA indicates the knowledge-based QA task.} \\
\multicolumn{4}{l}{$^{\mathrm{b}}$ MR-QA denotes the mathematical reasoning QA task.}
\end{tabular}
\label{tab_1}
\end{center}
\end{table}

\subsection{Model training settings}

We fine-tuned the Fino1 model~\cite{qian2025fino1transferabilityreasoningenhanced} for our DRR task using supervised instruction tuning. The model was trained with a block size of 4096 tokens and a maximum context length of 8192 tokens, enabling it to handle long-form document reasoning. To enable efficient training on limited GPU memory, we applied parameter-efficient fine-tuning (PEFT) via LoRA, with rank $r=64$, scaling factor $\alpha=128$, and a dropout rate of 0.05. We enabled int4 quantization and applied LoRA to all linear modules.

The training was conducted for 10 epochs with a batch size of 1, using gradient accumulation over 4 steps to simulate a larger effective batch size. We adopted the AdamW optimizer with a learning rate of 3e-5, a cosine learning rate scheduler, and a warmup ratio of 1\%. Training was performed with bf16 mixed-precision on 4 NVIDIA H100 GPUs.

\section{Experiment and result}

\subsection{Evaluation dataset}
To better evaluate our RKEFino1 model, we used evaluation data from the regulation challenge of FinNLP-FNP-LLMFinLegal-2025 shared task\footnote{\url{https://github.com/Open-Finance-Lab/Regulations_Challenge_COLING_2025}}~\cite{wang2025finnlp}, with the statistical details summarized in Table~\ref{tab_2}. For the Knowledge-based QA (K QA) task, we constructed 150 test samples, and for the Mathematical Reasoning QA (MR QA) task, we collected 50 test samples. To assess the model's generalizability, we introduced a new task called Numerical NER, inspired by FiNER~\cite{loukas-etal-2022-finer} and FNXL~\cite{sharma2023financial}. Unlike previous work, our Numerical NER task focuses on identifying five entity types (Integer Item Type, Monetary Item Type, Per Share Item Type, Percent Item Type, and Shares Item Type) instead of us gaap tags, and it processes both sentences and tables rather than sentences only.

\begin{table}[htbp]
\caption{The statistic of evaluation data.}
\begin{center}
\begin{tabular}{|c|c|c|c|}
\hline
\textbf{Task} & \textbf{Domain}& \textbf{\#Samples}& \textbf{Source} \\
\hline
\multirow{3}*{K-QA} & CDM& 126 & CDM documentation \\
~& MOF& 161 & OSI website \\
~ & XBRL& 700 & SEC and XBRL Terminology \\
\hline
MR-QA & XBRL & 1,000 & SEC \\
\hline
Numerical NER & XBRL & 3,638 & SEC \\
\hline
\end{tabular}
\label{tab_2}
\end{center}
\end{table}

\begin{table*}[htbp]
\caption{Comparison of Fino1 and RKEFino1 on fine-grained tasks.}
\begin{center}
\begin{tabular}{|c|c|c|c|c|}
\hline
\textbf{Task} & \textbf{Fine-grained Task} & \textbf{Metrics} & \textbf{Fino1} & \textbf{RKEFino1}
\\
\hline
\multirow{7}*{K-QA} 
& CDM-QA           &  FActScore (\%)  & 36.76 & 42.58
\\
& MOF Abbreviation &  Accuracy (\%) & 0.00 & 12.23
\\
& MOF Approval     & Accuracy (\%) & 0.00 & 62.58
\\
& MOF Details      & FActScore (\%) & 27.13 & 40.56
\\
& XBRL Domain      & FActScore (\%) & 20.08 & 45.87
\\
& XBRL Tag         & Accuracy (\%) & 0.00 & 16.02
\\
& XBRL Term        & FActScore (\%) & 26.22 & 50.28
\\
\hline
MR-QA 
& XBRL Math        & Accuracy (\%) & 56.87 & 70.69
\\
\hline
NER 
& Numerical NER    & F1-score (\%) & 14.99 & 26.62
\\
\hline
\end{tabular}
\label{tab_3}
\end{center}
\end{table*}

\subsection{Evaluation metrics}
To keep consistent with the competition, we also leveraged Accuracy and FactScore to evaluate our model's performance on K-QA and MR-QA tasks. Furthermore, we used F1-scores to evaluate performance on the numerical NER task.

\begin{itemize}
    \item Accuracy is primarily used for questions that require precise answers, such as full expansions of abbreviations, yes-or-no questions, and financial mathematical reasoning.
    \item FactScore is mainly used for question-answering scenarios, including CDM documentation, MOF detailed QA, and explanations of XBRL terms.
    \item F1-scores, commonly used in named entity recognition, is the harmonic mean of precision and recall, reflecting overall performance and generalization.
    
\end{itemize}


\subsection{Initial results}
To quantify the effectiveness of domain knowledge integration, we conduct a detailed comparison between the baseline Fino1 model and its enhanced variant RKEFino1, which is trained with additional knowledge. The results are presented in Table~\ref{tab_3}.



RKEFino1 significantly outperforms Fino1 across all three evaluation tasks. In knowledge-based QA, it achieves much higher accuracy on MOF Approval (62.58\% vs. 0.00\%) and XBRL Tag (16.02\% vs. 0.00\%), and shows notable FactScore gains on CDM-QA, MOF Details, and XBRL Term, highlighting the benefits of knowledge injection for factual correctness and relevance.

In the MR-QA task, which involves mathematical reasoning over financial data, RKEFino1 achieves 70.69\% accuracy, clearly surpassing Fino1’s 56.87\%, suggesting better grounding through domain knowledge.

For the newly proposed Numerical NER task, RKEFino1 reaches an F1-score of 26.62\%, compared to Fino1’s 14.99\%, demonstrating improved generalization and structured information extraction across both textual and tabular inputs.

\section{Conclusion and Future Work}
We present RKEFino1, an enhanced version of Fino1 that integrates regulatory knowledge to better address challenges in DRR. Fine-tuned on curated data from XBRL, CDM, and MOF, it performs domain-specific reasoning via supervised learning guided by a verifier.
To evaluate its capabilities, we design two QA tasks—one on regulatory knowledge, the other on mathematical reasoning—and a numerical NER task to test generalization across regulatory data.
Experiments show RKEFino1 significantly outperforms Fino1 in all DRR tasks and excels at recognizing numerical entities.
In future work, we plan to expand the dataset for MOF abbreviation and XBRL tag to further improve the model’s performance on this task.




\bibliographystyle{IEEEtran}
\bibliography{custom}

\end{document}